\documentclass[conference]{IEEEtran}
\IEEEoverridecommandlockouts
\usepackage{cite}
\usepackage{amsmath,amssymb,amsfonts}
\usepackage{graphicx}
\usepackage{textcomp}
\usepackage{xcolor}
\usepackage{booktabs}
\usepackage{amsthm}
\usepackage{url}
\usepackage{tikz}
\usepackage{pgfplots}
\pgfplotsset{compat=1.18}
\usepgfplotslibrary{groupplots}
\usetikzlibrary{arrows.meta,positioning,fit}
\newtheorem{proposition}{Proposition}
\newcommand{\tablebodyfont}{\fontsize{7.4}{8.1}\selectfont}
\def\BibTeX{{\rm B\kern-.05em{\sc i\kern-.025em b}\kern-.08em
    T\kern-.1667em\lower.7ex\hbox{E}\kern-.125emX}}
\begin{document}

\title{When Successful Knowledge Graph Edits Displace Correct Answers: Rank-Level Locality beyond Parameter Support}

\author{\IEEEauthorblockN{Yi-Cheng Lai}
\IEEEauthorblockA{\textit{Institute of Information Science, Academia Sinica}\\
Taipei, Taiwan \\
laiyicheng0@iis.sinica.edu.tw}
\and
\IEEEauthorblockN{Jerry Wang}
\IEEEauthorblockA{\textit{University of Illinois Urbana-Champaign}\\
Urbana, IL, USA \\
jerryw10@illinois.edu}
\and
\IEEEauthorblockN{Hsin-Ling Hsu}
\IEEEauthorblockA{\textit{National Chengchi University}\\
Taipei, Taiwan \\
112306092@nccu.edu.tw}
\and
\IEEEauthorblockN{Li-Chu Chi}
\IEEEauthorblockA{\textit{National Chengchi University}\\
Taipei, Taiwan \\
113306077@g.nccu.edu.tw}
\and
\IEEEauthorblockN{Ya-Wen Teng}
\IEEEauthorblockA{\textit{National Chengchi University}\\
Taipei, Taiwan \\
ywteng@nccu.edu.tw}
\and
\IEEEauthorblockN{Hen-Hsen Huang}
\IEEEauthorblockA{\textit{Institute of Information Science, Academia Sinica}\\
Taipei, Taiwan \\
hhhuang@iis.sinica.edu.tw}
}

\maketitle

\begin{abstract}
Editing a knowledge graph embedding (KGE) model to promote a desired answer
can displace correct answers from the returned list. Locality tests based only
on facts that reuse the edited parameter can miss this ranking effect. We
introduce a common rank-displacement audit at three scopes: facts supported by
the edited parameter, other correct answers to the target query, and correct
answers across queries with the same relation. We also derive dimensional and
geometric conditions for an update to improve the target while exactly
preserving selected scores. On FB15k-237 with DistMult and ComplEx, direct
promotion always moves the target into the top ten, but does so without damage
in only 23.0--23.2\% of edits. Strict preservation causes no measured damage,
yet succeeds in only 1.3--1.4\%. Support-regularized entity editing gives the
highest joint success, 36.3--37.7\%, while rank-truncated preservation reaches
32.8--34.7\% and reduces the mean number of displaced answers from about 14 to
1.2. Experiments across dimensions, scorers, ranking conventions, and a learned
editor show that locality depends on both the protected scope and the editing
mechanism. KGE editing should therefore report correction success together with
the incidence and severity of rank displacement.
\end{abstract}


\section{Introduction}
Knowledge graph embedding (KGE) models score candidate facts using learned representations of entities and relations. When a model ranks a desired fact too low, an edit can change a small set of parameters to promote it without retraining the model. A useful correction should also preserve other answers. For ranking models, however, these two goals are coupled: raising one candidate can push another below the top-$k$ cutoff even when the latter's score stays unchanged.

\begin{figure*}[t]
\centering

\definecolor{cTarget}{HTML}{3A6FD0}
\definecolor{cPos}{HTML}{2E7D4F}
\definecolor{cLost}{HTML}{C0453B}
\definecolor{cInk}{HTML}{2B2B2B}

\begin{tikzpicture}[
 font=\sffamily\scriptsize,
 x=1cm,
 y=1cm,
 sec/.style={
   font=\sffamily\footnotesize\bfseries,
   text=cInk,
   anchor=west
 },
 sub/.style={
   font=\sffamily\tiny,
   text=black!50,
   anchor=west
 },
 rowlab/.style={
   font=\sffamily\tiny,
   text=black!65,
   anchor=east
 },
 note/.style={
   font=\sffamily\tiny,
   text=black!50,
   anchor=west
 },
 chip/.style={
   rounded corners=1.5pt,
   draw=black!22,
   fill=white,
   minimum width=1.45cm,
   minimum height=.42cm,
   inner sep=1pt,
   font=\sffamily\tiny,
   text=black!65
 },
 dot/.style={
   circle,
   minimum size=4.6mm,
   inner sep=0pt,
   draw=none
 },
 other/.style={dot,fill=black!20},
 tgt/.style={dot,fill=cTarget},
 kept/.style={dot,fill=cPos},
 lost/.style={dot,fill=cLost},
 hair/.style={
   draw=black!40,
   line width=.35pt
 },
 lead/.style={
   -{Latex[length=1.3mm]},
   draw=black!35,
   line width=.4pt
 }
]

\def\LX{0}
\def\LW{4.30}
\def\RX{5.20}
\def\RW{10.80}
\def\ytop{3.60}

\def\rA{2.62}
\def\rB{1.86}
\def\rC{1.10}

\def\dA{7.75}
\def\dB{8.43}
\def\dC{9.11}
\def\dD{9.79}
\def\cut{9.45}

\node[sec] at (\LX,\ytop+.62) {Parameter support};
\node[sec] at (\RX,\ytop+.62) {Rank-level locality};

\draw[hair]
  (\LX,\ytop+.40) -- (\LX+\LW,\ytop+.40);
\draw[hair]
  (\RX,\ytop+.40) -- (\RX+\RW,\ytop+.40);

\node[sub] at (\LX,\ytop+.16) {direct score dependence};
\node[sub] at (\RX,\ytop+.16)
  {competition within ranked answer lists};

\node[
  circle,
  draw=cTarget,
  fill=cTarget!8,
  text=cTarget,
  line width=.6pt,
  minimum size=1.30cm,
  align=center,
  font=\sffamily\tiny\bfseries
] (emb) at (.90,\rB+.06) {edited\\vector};

\node[chip] (fa) at (3.15,\rA+.02) {fact};
\node[chip] (fb) at (3.15,\rB+.06) {fact};
\node[chip] (fc) at (3.15,\rC+.10) {fact};

\draw[lead] (emb) -- (fa);
\draw[lead] (emb) -- (fb);
\draw[lead] (emb) -- (fc);

\node[
  sub,
  text width=4.25cm,
  align=left
] at (\LX,.34) {
  Facts whose own scores reuse the edited vector.
  Their scores change directly.
};

\node[
  font=\sffamily\scriptsize\bfseries,
  text=cPos,
  anchor=east
] at (\RX+\RW-.18,2.96) {Relation-scope locality};
\filldraw[
  rounded corners=3pt,
  draw=cPos!70,
  fill=cPos!3,
  line width=.5pt
]
  (\RX+.02,.50)
  rectangle
  (\RX+\RW,3.20);

\node[
  font=\sffamily\tiny\bfseries,
  text=cPos,
  anchor=east
] at (\RX+\RW-.18,2.98) {relation scope};


\node[
  font=\sffamily\scriptsize\bfseries,
  text=cTarget,
  anchor=west
] at (\RX+.40,2.82) {Same-query locality};

\filldraw[
  rounded corners=2pt,
  draw=cTarget!70,
  fill=cTarget!4,
  line width=.5pt
]
  (\RX+.22,1.40)
  rectangle
  (\RX+5.35,3.06);

\node[
  font=\sffamily\tiny\bfseries,
  text=cTarget,
  anchor=west
] at (\RX+.40,2.84) {same query};

\draw[
  -{Latex[length=1.4mm]},
  draw=cTarget!80,
  line width=.6pt
]
  (\dD,2.82) -- (\dA,2.82);

\draw[
  -{Latex[length=1.4mm]},
  draw=cLost!80,
  line width=.6pt
]
  (\dB,1.56) -- (\dD,1.56);

\node[rowlab] at (\dA-.60,2.48) {before};
\node[other] at (\dA,2.48) {};
\node[kept]  at (\dB,2.48) {};
\node[other] at (\dC,2.48) {};
\node[tgt]   at (\dD,2.48) {};

\node[rowlab] at (\dA-.60,1.92) {after};
\node[tgt]   at (\dA,1.92) {};
\node[other] at (\dB,1.92) {};
\node[other] at (\dC,1.92) {};
\node[lost]  at (\dD,1.92) {};

\draw[
  densely dashed,
  black!40,
  line width=.4pt
]
  (\cut,1.66) -- (\cut,2.72);

\node[note,text=cLost] at (\RX+5.72,2.18) {
  target enters; a correct answer leaves
};

\node[rowlab] at (\dA-.60,.96) {other query};
\node[other] at (\dA,.96) {};
\node[kept]  at (\dB,.96) {};
\node[other] at (\dC,.96) {};
\node[tgt]   at (\dD,.96) {};

\draw[
  densely dashed,
  black!40,
  line width=.4pt
]
  (\cut,.66) -- (\cut,1.26);

\node[note] at (\RX+5.72,.96) {
  the same competition can recur
};

\node[tgt] at (\RX+.12,-.46) {};
\node[note] at (\RX+.36,-.46) {edited target};

\node[kept] at (\RX+2.10,-.46) {};
\node[note] at (\RX+2.34,-.46) {retained positive};

\node[lost] at (\RX+4.55,-.46) {};
\node[note] at (\RX+4.79,-.46) {displaced positive};

\node[other] at (\RX+7.15,-.46) {};
\node[note] at (\RX+7.39,-.46) {other candidate};

\end{tikzpicture}

\caption{Three locality scopes. Dashed lines mark the top-$k$ cutoff.}
\label{fig:concept}
\end{figure*}
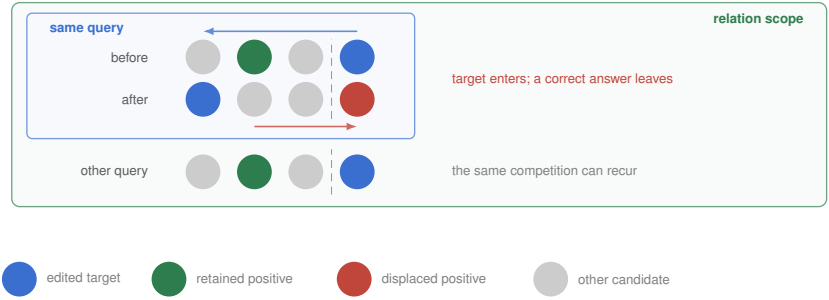

Consider a service that answers ``what is this person's occupation?''
Suppose the model ranks \emph{chemist} too low for Marie Curie, and we
edit the embedding of \emph{chemist} to promote it. Three different sets
of answers are now at risk. First, other facts that use the same edited
vector, such as the occupation of Linus Pauling: this is the set a
parameter-support audit checks. Second, Curie's own remaining occupations. Suppose \emph{physicist} sits
at rank ten. Promoting \emph{chemist} past it pushes \emph{physicist} to
rank eleven and out of the returned answers, even though the score of
\emph{physicist} was never touched: what changed is the number of
candidates ahead of it.
Third, the answers returned for everyone else. The edit changes the
embedding of \emph{chemist}, which enters the score of every occupation
query, so \emph{chemist} rises for Einstein and Pauling as well. Wherever
a correct answer sits at the cutoff, it is displaced. We call
these \emph{parameter-support}, \emph{same-query}, and \emph{relation-scope}
locality. Only the first is defined by parameter sharing; the other two
are defined by competition within a ranked list, and an edit can leave the
first intact while damaging the other two. Figure~\ref{fig:concept} illustrates these three scopes.

We evaluate these three scopes using a common damage event: a known positive that was in the top ten before an edit falls below that cutoff afterward. This separates the choice of protected facts from the statistic used to measure damage. We then examine how editing strategies trade successful correction against answer preservation. Adapting AlphaEdit's null-space construction~\cite{alphaedit}, we use standard rank--nullity and projection tests to determine whether an exact score-preserving update can promote the target. Partial preservation relaxes these constraints, while competitor demotion attempts to improve the target's rank by lowering other candidates' scores.

\textbf{Contributions.} We (i) introduce a common rank-displacement audit that separates parameter-support, same-query, and relation-scope locality; (ii) derive dimensional and geometric tests for the feasibility of exact score-preserving edits; and (iii) compare promotion, preservation, demotion, and fine-tuning routes under matched edits. The experiments identify when partial preservation is useful, when exact preservation becomes too restrictive, and how blocker selection and model geometry govern collateral damage. A KGEditor case study extends the audit to a learned editor.

\section{Related Work}
\textbf{KGE and update settings.} TransE~\cite{transe} scores a relation as a vector offset; RESCAL~\cite{rescal}, DistMult~\cite{distmult}, and ComplEx~\cite{complex} are bilinear semantic-matching models; RotatE~\cite{rotate} represents relations as complex rotations. KGE unlearning and continual learning instead remove facts or retain aggregate performance across snapshots~\cite{fedlu,dicgrl,lkge,incde,temporal}. We study a fixed checkpoint, one supplied correction, and case-level rank displacement.

\textbf{Editing and locality.} KGEditor~\cite{kgeditor} studies fact editing and addition in a pretrained-language-model-based KGE and evaluates locality through its reference-triple metric RK@$k$. KnowledgeEditor (KE)~\cite{ke} and Model Editor Networks using Gradient Decomposition (MEND)~\cite{mend} learn parameter updates and assess changes outside the edited association. Ripple-effect and knowledge-conflict evaluations instead test whether edits propagate consistently to related facts~\cite{cohen-ripple,li-pitfalls}. We isolate a different effect: an unchanged correct answer can leave the top-$k$ when a competitor moves above it. Parameter-support locality follows the modified vector, relation scope follows a query family, and same-query locality follows one answer list.

\textbf{Projection-based control.} Orthogonal Gradient Descent, Gradient Projection Memory, and Averaged Gradient Episodic Memory project updates using protected directions~\cite{ogd,gpm,agem}; Elastic Weight Consolidation penalizes changes to important parameters~\cite{ewc}. ROME and MEMIT apply structured updates to language models~\cite{rome,memit}, and AlphaEdit projects updates into the null space of preserved knowledge~\cite{alphaedit}. In a linear KGE parameterization, selected score sensitivities form an explicit matrix for each edit. Its rank and target projection characterize exact score preservation, while post-edit ranking measures whether the selected answers remain in the top-$k$.

\section{Preliminaries}
\label{sec:prelim}
\subsection{Scores and Candidate Rankings}
We write a fact as $(h,r,t)$, where $h$ is the head entity, $r$ the relation, and $t$ the tail entity. For a query $(h,r,?)$, the model assigns each candidate tail a score $s(h,r,t)$ and ranks candidates from highest to lowest. Training encourages known facts to receive higher scores than negative examples.

Our experiments cover five common KGE scorers. DistMult, ComplEx, and RESCAL are linear in a tail embedding when the head and relation are fixed. RotatE and TransE instead use distance-based scores. This distinction matters because a vector update changes the first group of scores exactly linearly; for the distance-based models, the same calculation is only a local approximation. In every edit, all parameters outside the stated edited vectors remain fixed.

\subsection{Editing as a Perturbation}
An edit adds an update vector $\delta$ to a selected entity embedding. Let $s'$ denote the score after the edit. When the score is linear in the edited embedding, its change is exactly
\begin{equation}
s'(h,r,t)-s(h,r,t)=\langle \delta,c\rangle .
\label{eq:core}
\end{equation}
The sensitivity vector $c$ describes how that score responds to the edit. An update orthogonal to $c$ leaves the score unchanged. For RotatE and TransE, we use the gradient as $c$, so Eq.~\eqref{eq:core} is a first-order approximation. Our exact-preservation analysis therefore applies only when the score is linear in the edited embedding.

Suppose the edit targets the fact $(h^*,r^*,t^*)$ by promoting $t^*$ for the target query $(h^*,r^*,?)$. For a different protected query $(h,r,?)$, the same entity $t^*$ is also a candidate answer. Updating the embedding of $t^*$ may therefore change its score $s(h,r,t^*)$ in the protected query, potentially lowering the rank of the query's original correct answer.

\subsection{Edit Routes}
For this edit, let $\tau_k$ denote the score of the current $k$th-ranked tail, and let $\epsilon$ be a small positive margin. The score increase required to move the target strictly above the top-$k$ cutoff is
\[
\Delta=\max\{0,\tau_k-s(h^*,r^*,t^*)+\epsilon\}.
\]
Let $e_t$ denote the embedding of entity $t$. For the target fact, we write $g=\nabla_{e_{t^*}}s(h^*,r^*,t^*)$ for its sensitivity vector, corresponding to $c$ in Eq.~\eqref{eq:core}.

\textbf{Direct target promotion} uses the minimum-norm update that provides the required score increase:
\[
\delta=\frac{\Delta}{\lVert g\rVert_2^2}g.
\]
For a linear-tail scorer, this places the target in the top-$k$ without constraining other scores. For RotatE, we instead move the tail toward the rotated-head point and use binary search for a minimally sufficient interpolation coefficient.

\textbf{Support-regularized entity editing} balances the required target increase against changes to known facts that use the same tail:
\[
\min_\delta (g^\top\delta-\Delta)^2+\lVert C_{\rm ent}\delta\rVert_2^2 .
\]
$C_{\rm ent}$ contains sensitivity rows for known facts whose tail is the edited entity $t^*$. Because target promotion is penalized rather than enforced as a hard constraint, this method does not guarantee that the target reaches the top-$k$.

\textbf{Strict relation-nullspace promotion} protects training queries with relation $r^*$ whose correct answers have a pre-edit rank at most $k$. For each protected query $(h,r^*,?)$, it constrains the competing score $s(h,r^*,t^*)$ of the edited tail to remain unchanged. Stacking the corresponding sensitivity rows gives the constraint matrix $R$. The projector $P=I-R^\dagger R$, where $R^\dagger$ is the Moore--Penrose pseudoinverse, removes every component that would change a protected score. We project the target direction as $u=Pg$ and apply
\[
\delta=\frac{\Delta}{g^\top u}u
\]
provided that $u\neq 0$. This update exactly preserves the constrained competing scores. If the projection removes the entire target direction, the method returns a zero update.

\textbf{Rank-truncated preservation} relaxes strict preservation. Let $R=U\Sigma V^\top$, and let $V_m$ contain the first $m$ columns of $V$. We define
\[
P_m=I-V_mV_m^\top
\]
and use $u_m=P_mg$ in the promotion update above. Increasing $m$ removes more constraint directions from the update. We index this strength by $\rho=m/d$, where $d$ is the embedding dimension, and define the endpoints as direct promotion at $\rho=0$ and the full strict projector at $\rho=1$. For comparison, \textbf{ridge preservation} replaces the hard preservation constraint with a soft penalty:
\[
\min_\delta
(g^\top\delta-\Delta)^2
+\lambda\lVert R\delta\rVert_2^2 .
\]
The parameter $\lambda\geq 0$ controls the trade-off between target promotion and protected-score preservation.

\section{Locality at Three Scopes}

For all three scopes, we use raw tail rank, defined as one plus
the number of candidate tails with a strictly higher score; ties
are therefore not counted as ahead. Raw ranking retains competition
among known positives, whereas filtered ranking removes other known
positives and measures a different form of answer preservation.

An edit succeeds at $k$ if the target's post-edit raw tail rank is
at most $k$; S@$k$ denotes the proportion of successful edits.
A protected fact is damaged if its raw tail rank is at most $k$
before the edit but exceeds $k$ afterward. The three locality
scopes use this same damage event but differ in the set of facts
they protect.

\textbf{Parameter-support locality} evaluates known facts whose
scores directly depend on an edited parameter. Because the edited
parameters differ across edit routes, the corresponding protected
set is defined separately for each route. This scope captures
collateral damage caused by direct parameter reuse.

\textbf{Same-query locality} evaluates the other known-positive
tails of the target query $(h^*,r^*,?)$, excluding the edit target.
It captures rank displacement within the edited answer list.

\textbf{Relation-scope locality} extends the audit from the target
query to known answers of other queries that share the edited
relation, including queries with different heads. It measures
whether an edit disrupts answer lists across the relation rather
than only within the target query.

For each edit, $D$ counts the protected facts that fall out of
the top-$k$. Across all evaluated edits, S@$k$ reports the editing
success rate, ND reports the rate of edits with no damage, and
SafeSucc reports the rate of edits that both succeed and cause no
damage. The mean, 95th percentile (P95), and maximum of $D$
describe damage severity.

\section{Feasibility of Exact Score Preservation}
\label{sec:walls}
Exact score preservation asks the edit to improve the target while leaving every selected constraint score unchanged. This is stronger than preserving top-$k$ membership. As in Sec.~\ref{sec:prelim}, let $R$ contain one sensitivity row per constraint. Its null space, $\mathcal{S}=\ker(R)$, contains exactly the update directions that preserve all selected scores. Two tests determine whether this space can support the edit.

\subsection{Constraint Rank}

By Eq.~\eqref{eq:core}, preserving all selected scores requires
$R\delta=0$. The feasible updates therefore lie in the null space
$\mathcal{S}=\ker(R)$.

\begin{proposition}[Dimensional Wall]
The preserving subspace has dimension
\[
\dim\mathcal{S}=d-\operatorname{rank}(R).
\]
A nonzero score-preserving update exists if and only if
$\operatorname{rank}(R)<d$.
\par\smallskip\noindent\upshape\emph{Proof.}
By rank--nullity,
$\dim\ker(R)=d-\operatorname{rank}(R)$; hence the null space
contains a nonzero vector exactly when $\operatorname{rank}(R)<d$.
\hfill$\square$
\end{proposition}

Each linearly independent sensitivity row increases the rank of
$R$ and removes one degree of freedom from the preserving
subspace. In the primary $d=256$ setting, a relation can provide
thousands of constraints. Among the 16 evaluated relations with
at least 1,000 protected facts, 15 produce
$\operatorname{rank}(R)=d$, and the mean null-space dimension is
only $1.125$. Exact score-preserving updates therefore have little
or no room for most of these relations. When $\operatorname{rank}(R)=d$, the zero vector is the only
score-preserving update and cannot increase the target score. 

\subsection{Target Alignment}

A nonzero preserving direction is useful only if it can also
increase the target score. Let $\Pi_{\mathcal S}$ denote the
orthogonal projector onto the preserving subspace $\mathcal S$.
The component $\Pi_{\mathcal S}g$ measures how much of the target
sensitivity remains available after imposing the preservation
constraints.

\begin{proposition}[Geometric Wall]
Assume $g\neq0$, and define the target-alignment coefficient
\[
E=\frac{\lVert\Pi_{\mathcal S}g\rVert}
        {\lVert g\rVert}\in[0,1].
\]
Under an update-norm bound $\lVert\delta\rVert\leq B$,
\begin{equation}
\max_{\delta\in\mathcal S,\,\lVert\delta\rVert\leq B}
\langle\delta,g\rangle
=
B\lVert\Pi_{\mathcal S}g\rVert
=
B\lVert g\rVert E.
\end{equation}
In particular, if $E=0$, no score-preserving update can increase
the target score within this parameterization.
\par\smallskip\noindent\upshape\emph{Proof.}
For $\delta\in\mathcal S$,
$\langle\delta,g\rangle
=\langle\delta,\Pi_{\mathcal S}g\rangle$.
By Cauchy--Schwarz, this is at most
$B\lVert\Pi_{\mathcal S}g\rVert$. When
$\Pi_{\mathcal S}g\neq0$, equality is attained by
\[
\delta
=
B\frac{\Pi_{\mathcal S}g}
        {\lVert\Pi_{\mathcal S}g\rVert}.
\]
If $\Pi_{\mathcal S}g=0$, the maximum is zero.
\hfill$\square$
\end{proposition}

Together, the two propositions give two feasibility tests. Without
a norm bound, exact promotion requires both
$\operatorname{rank}(R)<d$ and $E>0$: a preserving direction must
exist, and it must retain a component aligned with the target.
Under the additional bound $\lVert\delta\rVert\leq B$, promotion
also requires
$B\lVert g\rVert E\geq\Delta$.

\textbf{Empirical coverage.}
Across all eligible FB15k-237 targets, only 191 of 11,339 DistMult
cases ($1.68\%$) and 185 of 11,305 ComplEx cases ($1.64\%$) pass
both tests. For DistMult, 6,410 cases fail the rank test and 4,738
of the remaining cases fail the alignment test; the corresponding
ComplEx counts are 6,398 and 4,722. Every passing case produces a
nonzero update that reaches the top-$k$ without observed
relation-scope damage (191/191 and 185/185). The 95\%
Clopper--Pearson lower bounds for these conditional success rates
are $.981$ and $.980$. Rejected cases receive a zero update by
design, so the pass rate measures the coverage of exact
preservation.

\textbf{From score preservation to rank safety.}
Exact score preservation and rank safety are not generally
equivalent. Holding a protected answer's own score fixed does not
preserve its rank if a competitor moves, whereas some score changes
do not alter top-$k$ membership. Our strict target-tail route instead
constrains the edited tail's competing score in each protected query.
We nevertheless evaluate post-edit ranks as the final measure of
answer preservation. Because exact promotion covers few targets, we
next consider improving the target's rank by lowering its competitors.

\section{Competitor Demotion with Observed Labels}
\label{sec:demotion}
Instead of raising the target's score, an edit can lower competing candidates. We use this intervention to measure how often rank correction is possible through parameters other than the target tail.

\textbf{Observed-unlabeled blockers.}
For an edit query $(h,r,t^*)$, any tail $t_b$ satisfying
$s(h,r,t_b)>s(h,r,t^*)$ is a \emph{blocker}. We call it
\emph{observed-unlabeled} if $t_b\notin A(h,r)$, where $A(h,r)$
contains the known-positive tails in
$\mathrm{train}\cup\mathrm{valid}\cup\mathrm{test}$.

\textbf{Confidence filter.}
Let $s_{\max}$ be the highest candidate score for the edit query.
We retain only blockers satisfying
\[
s(h,r,t_b)
\leq
s(h,r,t^*)+q\bigl(s_{\max}-s(h,r,t^*)\bigr),
\]
where $q\in[0,1]$ controls the filtering strength. Setting $q=1$
retains all blockers, while smaller values exclude more
high-scoring blockers.

\textbf{Closed-form bilinear demotion.}
This route modifies the query head,
$e_h\leftarrow e_h+\delta_h$, while keeping the target-tail
embedding fixed. If the target currently has rank
$\operatorname{rank}(t^*)$, at least
\[
n_{\rm req}
=
\max\{\operatorname{rank}(t^*)-k,0\}
\]
candidates must move below it for the target to enter the top-$k$.

Let $\mathcal B=\{b_1,\ldots,b_{n_{\rm req}}\}$ be the selected
observed-unlabeled blockers, and write
$s_x=s(h,r,x)$. For each candidate $x$, define its sensitivity
to the head update as
\[
a_x=\nabla_{e_h}s(h,r,x).
\]
The desired change for blocker $b$ is
$s_{t^*}-s_b-\epsilon$, which places it just below the unchanged
target score. Stacking these requests gives
\begin{equation}
\begin{aligned}
A_{\mathcal B}
&=
\begin{bmatrix}
a_{b_1}^{\top}\\
\vdots\\
a_{b_{n_{\rm req}}}^{\top}\\
a_{t^*}^{\top}
\end{bmatrix},
&
y_{\mathcal B}
&=
\begin{bmatrix}
s_{t^*}-s_{b_1}-\epsilon\\
\vdots\\
s_{t^*}-s_{b_{n_{\rm req}}}-\epsilon\\
0
\end{bmatrix},\\
\delta_h
&=
\arg\min_{\delta}
\lVert A_{\mathcal B}\delta-y_{\mathcal B}\rVert_2^2 .
\end{aligned}
\label{eq:cf-demotion}
\end{equation}

The last row requests no change to the target score. Because one
head update must satisfy all rows simultaneously, the least-squares
solution may not achieve every requested score change exactly.

\textbf{Gradient-based blocker demotion.}
Unlike the closed-form route, this route updates the tail
embedding of each selected blocker:
$e_{t_b}\leftarrow e_{t_b}+\delta_b$. We optimize the pairwise
margin objective
\begin{equation}
\begin{aligned}
\mathcal{L}
={}&
\sum_{b\in\mathcal{B}}
\operatorname{softplus}\!\left(
s'(h,r,t_b)-s(h,r,t^*)+\mu
\right) \\
&+
\lambda
\sum_{b\in\mathcal{B}}
\lVert\delta_b\rVert_2^2,
\end{aligned}
\label{eq:gradient-demotion}
\end{equation}
subject to $\lVert\delta_b\rVert_2\leq B$ for every
$b\in\mathcal B$. Here, $s'(h,r,t_b)$ is evaluated using the
updated blocker embedding. The first term encourages each blocker
to fall below the target by margin $\mu$, while the second limits
the size of the updates. Because it uses automatic differentiation,
this route also applies when the closed-form bilinear construction
is unavailable.


\section{Experiments}
\label{sec:experiments}

\subsection{Setup}

\textbf{Evaluation cohorts.}
Our primary comparison uses public LibKGE DistMult and ComplEx
checkpoints on FB15k-237. We sample 1,000 deduplicated test triples per
model without replacement from targets ranked below the top 10; the
fixed cohorts span 110/118 relations. Separate 1,000-case cohorts with
rank above 20 support the $k\in\{5,10,20\}$ analysis. The norm study
also covers FB15k-237 RESCAL and five WN18RR scorers, with 1,000 edits
per scorer.

\textbf{Evaluation protocol.}
Unless stated otherwise, we evaluate raw tail rankings at $k=10$ and
recompute ranks from complete float64 score vectors. Correction success
(S@$k$) is the fraction of targets entering the top-$k$. Let $D$ count
protected answers that cross from rank $\leq k$ to rank $>k$. We report
$\Pr(D=0)$, SafeSucc (success with $D=0$), and the mean, 95th
percentile, and maximum of $D$.

\textbf{Protected sets.}
Relation-scope evaluation protects training facts whose pre-edit ranks
are at most $k$, capped at 5,000 per relation. Same-query evaluation
uses all-split known-positive tails for the target query, excluding the
target. Parameter-support evaluation uses training facts sharing the
edited tail, training facts sharing the edited head for closed-form
demotion, or all-split facts involving an edited blocker for gradient
demotion. Entity editing builds $C_{\rm ent}$ from at most 256
highest-norm all-split sensitivity rows sharing $t^*$.

\textbf{Implementation and model selection.}
Sampling and runs use seed 0 unless stated otherwise. We use
$\epsilon=10^{-6}$; RotatE direct promotion uses 32-step binary
search. Closed-form demotion uses one unbounded least-squares solve and
returns zero without $n_{\rm req}$ eligible blockers. Projected
denominators at most $10^{-12}$ are unusable, and updates below that
norm are identities. We evaluate eight rank-truncation settings
$\rho\in\{0,.125,.25,.5,.75,.875,.9375,1\}$, seven ridge penalties
$\lambda\in\{10^{-4},10^{-3},10^{-2},10^{-1},1,10,100\}$, and five
confidence thresholds $q\in\{0,.25,.5,.75,1\}$. Rank truncation,
ridge, fine-tuning, and replay use disjoint 200-case validation cohorts;
we maximize SafeSucc and break ties by S@10, no-damage rate, then the
smaller setting. Cases are paired within each sweep.

\textbf{Uncertainty.}
Confidence intervals use 5,000 bootstrap replicates over either edits
or relation clusters. Paired values below are reported in
DistMult/ComplEx order unless stated otherwise.

\textbf{Code availability.}
Code and evaluation artifacts are available at
\url{https://github.com/pa0lai/kge-rank-locality}.

\subsection{Main Results}
\label{sec:primary-pop}

\textbf{Dependence on locality scope.}
Table~\ref{tab:locality-scopes} gives opposite locality judgments for
the same correction depending on the protected scope. Direct promotion
causes no parameter-support damage in either model, yet damages
243/238 same-query cases and 262/259 relation-scope cases. The zeros do
not show that the rankings are globally preserved; they expose the blind
spot of checking only directly affected protected facts while the edited
tail moves as a competitor in other answer lists. Head-based demotion
reverses the pattern. Locality is therefore a property of the
editor--scope pair, not of the editor alone.

\begin{table}[t]
\centering
\tablebodyfont
\caption{Edits causing damage at each locality scope ($N=300$).}
\label{tab:locality-scopes}
\setlength{\tabcolsep}{3.2pt}
\begin{tabular}{lrrr}
\toprule
Editing route &
\shortstack{Parameter\\support} &
\shortstack{Same\\query} &
\shortstack{Relation\\scope} \\
\midrule
\multicolumn{4}{l}{\textit{DistMult}} \\
Direct promotion & 0 & 243 & 262 \\
Support-regularized entity editing & 21 & 221 & 233 \\
Closed-form demotion & 25 & 14 & 11 \\
\midrule
\multicolumn{4}{l}{\textit{ComplEx}} \\
Direct promotion & 0 & 238 & 259 \\
Support-regularized entity editing & 22 & 218 & 221 \\
Closed-form demotion & 34 & 22 & 18 \\
\bottomrule
\end{tabular}
\end{table}

\textbf{Overall comparison.}
The endpoints in Table~\ref{tab:primary} expose why success or safety
alone is insufficient: direct promotion always succeeds but is often
damaging, whereas strict preservation is safe largely because it rarely
edits. Entity editing gives the best joint outcome ($.363/.377$).
Rank truncation is close ($.328/.347$) while reducing mean displacement
from $14.942/13.943$ to $1.241/1.265$.

\begin{table*}[t]
\centering
\tablebodyfont
\caption{Primary FB15k-237 results ($N=1{,}000$ paired edits per model).}
\label{tab:primary}
\setlength{\tabcolsep}{4.0pt}
\begin{tabular}{llrrrrrr}
\toprule
Model & Editing route &
\shortstack{Target in\\top 10} &
\shortstack{No relation\\damage} &
\shortstack{Successful and\\no damage} &
\shortstack{Mean\\displaced} &
\shortstack{95th\\percentile} &
\shortstack{Maximum\\displaced} \\
\midrule
DistMult & Direct promotion                   & 1.000 & .230 & .230 & 14.942 & 101.00 & 295 \\
         & Support-regularized entity editing & .891 & .369 & \textbf{.363} & 3.410 & 22.05 & 142 \\
         & Rank-truncated preservation        & .912 & .416 & .328 & 1.241 & 4.05 & 22 \\
         & Strict null-space preservation     & .014 & 1.000 & .014 & 0 & 0 & 0 \\
         & Closed-form demotion               & .201 & .915 & .182 & .196 & 1.00 & 9 \\
         & Gradient-based demotion            & .129 & .793 & .031 & 21.920 & 78.25 & 2,631 \\
         & Naive target-tail fine-tuning      & .467 & .399 & .098 & 21.081 & 197.00 & 308 \\
         & Fine-tuning with replay            & .374 & .603 & .089 & 18.199 & 188.05 & 308 \\
\midrule
ComplEx  & Direct promotion                   & 1.000 & .232 & .232 & 13.943 & 90.10 & 266 \\
         & Support-regularized entity editing & .901 & .388 & \textbf{.377} & 2.971 & 12.00 & 195 \\
         & Rank-truncated preservation        & .904 & .443 & .347 & 1.265 & 5.00 & 33 \\
         & Strict null-space preservation     & .013 & 1.000 & .013 & 0 & 0 & 0 \\
         & Closed-form demotion               & .223 & .921 & .212 & .194 & 1.00 & 8 \\
         & Gradient-based demotion            & .135 & .771 & .029 & 24.159 & 117.05 & 1,525 \\
         & Naive target-tail fine-tuning      & .502 & .392 & .087 & 20.223 & 183.15 & 294 \\
         & Fine-tuning with replay            & .505 & .397 & .088 & 20.118 & 183.15 & 294 \\
\bottomrule
\end{tabular}
\end{table*}

Replay does not improve naive fine-tuning consistently ($.098$ to
$.089$ and $.087$ to $.088$). Entity editing exceeds rank truncation by
$.035$ for DistMult (relation-cluster 95\% CI $[.009,.065]$) and $.030$
for ComplEx ($[-.001,.063]$).

\textbf{Safety after a successful correction.}
The full-cohort no-damage rate also counts failed and identity edits.
Table~\ref{tab:conditional} therefore conditions on successful
corrections. Closed-form demotion is damage-free in $90.5/95.1\%$ of
successful cases, whereas gradient demotion displaces an average of
$89.171/66.170$ answers. The contrast shows that correction success
alone does not distinguish a controlled edit from one that crosses the
cutoff by broadly disturbing the ranking.

The failure modes also differ. Entity editing's mean displacement falls
from $3.410/2.971$ overall to $.637/.669$ among successful edits, showing
that its damage concentrates in failed corrections. Unsuccessful
rank-truncated edits cause no observed relation-scope displacement.

\begin{table*}[t]
\centering
\tablebodyfont
\caption{Relation-scope damage among successful corrections.}
\label{tab:conditional}
\begin{minipage}[t]{.49\textwidth}
\centering
\textit{(a) DistMult}\par\smallskip
\setlength{\tabcolsep}{3.2pt}
\begin{tabular}{lrrr}
\toprule
Editing route &
\shortstack{No-damage\\rate} &
\shortstack{Mean\\displaced} &
\shortstack{95th\\percentile} \\
\midrule
Direct promotion                   & .230 & 14.942 & 101.0 \\
Support-regularized entity editing & .407 & .637 & 1.0 \\
Rank-truncated preservation        & .360 & 1.361 & 5.0 \\
Strict null-space preservation     & 1.000 & 0 & 0 \\
Closed-form demotion               & .905 & .109 & 1.0 \\
Gradient-based demotion            & .240 & 89.171 & 202.8 \\
\bottomrule
\end{tabular}
\end{minipage}\hfill
\begin{minipage}[t]{.49\textwidth}
\centering
\textit{(b) ComplEx}\par\smallskip
\setlength{\tabcolsep}{3.2pt}
\begin{tabular}{lrrr}
\toprule
Editing route &
\shortstack{No-damage\\rate} &
\shortstack{Mean\\displaced} &
\shortstack{95th\\percentile} \\
\midrule
Direct promotion                   & .232 & 13.943 & 90.1 \\
Support-regularized entity editing & .418 & .669 & 1.0 \\
Rank-truncated preservation        & .384 & 1.399 & 5.0 \\
Strict null-space preservation     & 1.000 & 0 & 0 \\
Closed-form demotion               & .951 & .054 & 0 \\
Gradient-based demotion            & .215 & 66.170 & 266.9 \\
\bottomrule
\end{tabular}
\end{minipage}
\end{table*}

\subsection{Preservation Strength}
\label{sec:preservation-strength}

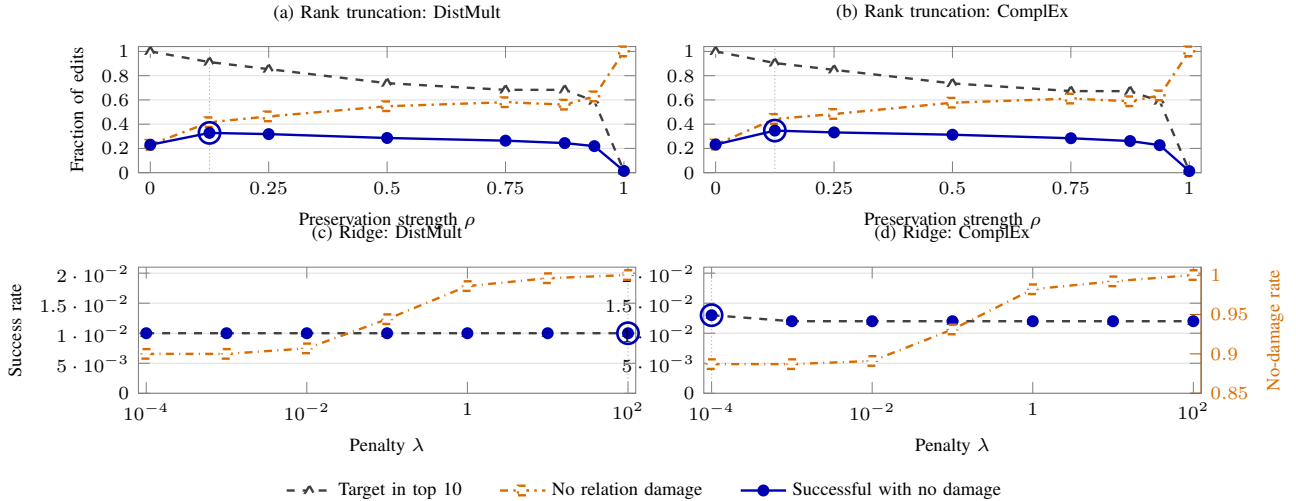
\begin{figure*}[t]
\centering
\begin{tikzpicture}
\begin{groupplot}[
 group style={group size=2 by 2,horizontal sep=.9cm,vertical sep=1.25cm},
 width=.45\textwidth,height=3.25cm,
 ymin=0,ymax=1.04,ytick={0,.2,.4,.6,.8,1},
 tick label style={font=\scriptsize},
 label style={font=\scriptsize},title style={font=\scriptsize},
 scaled ticks=false,ymajorgrids=true,grid style={black!12},
 axis line style={black!50},tick style={black!50},
 every axis plot/.append style={line width=.9pt,mark size=1.8pt},
 legend style={font=\scriptsize,draw=none,legend columns=3,
 /tikz/every even column/.append style={column sep=1.2em}},
]
\nextgroupplot[
 title={(a) Rank truncation: DistMult},
 xmin=-.025,xmax=1.025,xtick={0,.25,.5,.75,1},
 xlabel={Preservation strength $\rho$},
 ylabel={Fraction of edits},
 legend to name=preservationlegend]
\draw[black!35,densely dotted] (axis cs:.125,0)--(axis cs:.125,1.04);
\addplot[black!75,dashed,mark=triangle*,mark options={fill=white}]
 coordinates {(0,1.000) (.125,.912) (.25,.854) (.5,.738) (.75,.683) (.875,.683) (.9375,.590) (1,.014)};
\addlegendentry{Target in top 10}
\addplot[orange!85!black,dashdotted,mark=square*,mark options={fill=white}]
 coordinates {(0,.230) (.125,.416) (.25,.464) (.5,.548) (.75,.581) (.875,.561) (.9375,.629) (1,1.000)};
\addlegendentry{No relation damage}
\addplot[blue!70!black,solid,mark=*]
 coordinates {(0,.230) (.125,.328) (.25,.318) (.5,.286) (.75,.264) (.875,.244) (.9375,.219) (1,.014)};
\addlegendentry{Successful with no damage}
\addplot[only marks,blue!70!black,mark=o,mark size=4pt,line width=1pt,forget plot]
 coordinates {(.125,.328)};
\nextgroupplot[
 title={(b) Rank truncation: ComplEx},
 xmin=-.025,xmax=1.025,xtick={0,.25,.5,.75,1},
 xlabel={Preservation strength $\rho$}]
\draw[black!35,densely dotted] (axis cs:.125,0)--(axis cs:.125,1.04);
\addplot[black!75,dashed,mark=triangle*,mark options={fill=white}]
 coordinates {(0,1.000) (.125,.904) (.25,.848) (.5,.736) (.75,.673) (.875,.672) (.9375,.590) (1,.013)};
\addplot[orange!85!black,dashdotted,mark=square*,mark options={fill=white}]
 coordinates {(0,.232) (.125,.443) (.25,.484) (.5,.577) (.75,.611) (.875,.589) (.9375,.637) (1,1.000)};
\addplot[blue!70!black,solid,mark=*]
 coordinates {(0,.232) (.125,.347) (.25,.332) (.5,.313) (.75,.284) (.875,.261) (.9375,.227) (1,.013)};
\addplot[only marks,blue!70!black,mark=o,mark size=4pt,line width=1pt,forget plot]
 coordinates {(.125,.347)};
\nextgroupplot[
 title={(c) Ridge: DistMult},
 xmode=log,log basis x=10,xmin=8e-5,xmax=125,
 xtick={1e-4,1e-2,1,100},
 xticklabels={$10^{-4}$,$10^{-2}$,$1$,$10^2$},
 ymin=0,ymax=.021,ytick={0,.005,.01,.015,.02},
 xlabel={Penalty $\lambda$},ylabel={Success rate}]
\draw[black!35,densely dotted] (axis cs:100,0)--(axis cs:100,.021);
\addplot[black!75,dashed,mark=triangle*,mark options={fill=white}]
 coordinates {(1e-4,.010) (1e-3,.010) (1e-2,.010) (1e-1,.010) (1,.010) (10,.010) (100,.010)};
\addplot[blue!70!black,only marks,mark=*]
 coordinates {(1e-4,.010) (1e-3,.010) (1e-2,.010) (1e-1,.010) (1,.010) (10,.010) (100,.010)};
\addplot[only marks,blue!70!black,mark=o,mark size=4pt,line width=1pt,forget plot]
 coordinates {(100,.010)};
\nextgroupplot[
 title={(d) Ridge: ComplEx},
 xmode=log,log basis x=10,xmin=8e-5,xmax=125,
 xtick={1e-4,1e-2,1,100},
 xticklabels={$10^{-4}$,$10^{-2}$,$1$,$10^2$},
 ymin=0,ymax=.021,ytick={0,.005,.01,.015,.02},
 xlabel={Penalty $\lambda$}]
\draw[black!35,densely dotted] (axis cs:1e-4,0)--(axis cs:1e-4,.021);
\addplot[black!75,dashed,mark=triangle*,mark options={fill=white}]
 coordinates {(1e-4,.013) (1e-3,.012) (1e-2,.012) (1e-1,.012) (1,.012) (10,.012) (100,.012)};
\addplot[blue!70!black,only marks,mark=*]
 coordinates {(1e-4,.013) (1e-3,.012) (1e-2,.012) (1e-1,.012) (1,.012) (10,.012) (100,.012)};
\addplot[only marks,blue!70!black,mark=o,mark size=4pt,line width=1pt,forget plot]
 coordinates {(1e-4,.013)};
\end{groupplot}

\begin{axis}[
 at={(group c1r2.south west)},anchor=south west,
 width=.45\textwidth,height=3.25cm,
 xmode=log,log basis x=10,xmin=8e-5,xmax=125,
 axis x line=none,xtick=\empty,
 axis y line=none,ymin=.85,ymax=1.01,ytick=\empty,
 scaled ticks=false]
\addplot[orange!85!black,dashdotted,mark=square*,mark options={fill=white},line width=.9pt,mark size=1.8pt]
 coordinates {(1e-4,.900) (1e-3,.900) (1e-2,.907) (1e-1,.944) (1,.986) (10,.996) (100,1.000)};
\end{axis}
\begin{axis}[
 at={(group c2r2.south west)},anchor=south west,
 width=.45\textwidth,height=3.25cm,
 xmode=log,log basis x=10,xmin=8e-5,xmax=125,
 axis x line=none,xtick=\empty,
 axis y line*=right,ymin=.85,ymax=1.01,
 ytick={.85,.90,.95,1},
 ylabel={No-damage rate},
 tick label style={font=\scriptsize,text=orange!85!black},
 label style={font=\scriptsize,text=orange!85!black},
 axis line style={orange!70!black},tick style={orange!70!black},
 scaled ticks=false]
\addplot[orange!85!black,dashdotted,mark=square*,mark options={fill=white},line width=.9pt,mark size=1.8pt]
 coordinates {(1e-4,.887) (1e-3,.887) (1e-2,.891) (1e-1,.931) (1,.982) (10,.992) (100,1.000)};
\end{axis}
\end{tikzpicture}
\par\smallskip
\pgfplotslegendfromname{preservationlegend}
\caption{Preservation sweeps ($N=1{,}000$). Circles mark validation-selected settings; ridge success axes end at $2\%$.}
\label{fig:preservation-frontier}
\end{figure*}

Figure~\ref{fig:preservation-frontier} reveals a qualitative difference
between the two controls. Rank truncation creates a useful intermediate
region because it removes dominant preservation directions while still
retaining enough target-aligned capacity to cross the rank cutoff. Both
validation cohorts select $\rho=.125$, where test SafeSucc reaches
$.328/.347$. The wider relation-cluster intervals
($[.228,.451]/[.253,.454]$, versus case-bootstrap
$[.298,.358]/[.317,.378]$) indicate that the size of this gain depends
substantially on which relations are edited.

Stronger truncation is not uniformly safer. At
$\rho=.9375$, success falls to $.590$ for both models, mean displacement
rises to $2.461/2.376$, and the mean update norm grows to
$145.322/84.195$. The strict endpoint eliminates observed damage but
produces only 14/13 successful edits. Lower success together with the
larger update norm suggests that aggressive projection leaves a
poorly aligned direction: satisfying the remaining target objective
then requires a larger, not necessarily safer, parameter change.

Ridge does not exhibit the same intermediate region. Its high
no-damage rate is primarily a form of under-editing: increasing
$\lambda$ suppresses protected-score changes but also leaves target
success near $1\%$. Because every update is nonzero, the collapse in
editability cannot be attributed to identity updates; the soft penalty
instead fails to provide enough target movement to cross the discrete
rank cutoff. Validation selects $\lambda=100$ for DistMult and
$10^{-4}$ for ComplEx, with test SafeSucc $.010/.013$.

\subsection{Dimension and Checkpoint Geometry}

Table~\ref{tab:dimension} separates the two sources of infeasibility.
Larger embeddings reduce full-rank failure, but the remaining
subspaces align with the target less often; consequently, the fraction
passing both tests is non-monotonic. Across 18 training runs (two
models, three dimensions, and three seeds), rank truncation at
$d=512$ and $\rho=.125$ attains SafeSucc $.452\pm.015$ for DistMult
and $.500\pm.019$ for ComplEx.

\begin{table}[t]
\centering
\tablebodyfont
\caption{Exact-edit feasibility by embedding dimension ($N=1{,}000$).}
\label{tab:dimension}
\setlength{\tabcolsep}{2.7pt}
\begin{tabular}{lrrrrr}
\toprule
Model &
\shortstack{Dimen-\\sion} &
\shortstack{Full-rank\\failure} &
\shortstack{Alignment\\failure} &
\shortstack{Pass both\\tests} &
\shortstack{Safe exact\\edit} \\
\midrule
DistMult & 128 & .596 & .291 & .113 & .112 \\
         & 256 & .556 & .396 & .048 & .048 \\
         & 512 & .413 & .518 & .069 & .069 \\
ComplEx  & 128 & .633 & .274 & .093 & .093 \\
         & 256 & .531 & .397 & .072 & .072 \\
         & 512 & .353 & .479 & .168 & .168 \\
\bottomrule
\end{tabular}
\end{table}

Thus dimension alone does not determine editability; the learned score
geometry matters as well.

\subsection{Demotion Sensitivity}

\textbf{Update norm.}
Table~\ref{tab:norm-sensitivity} shows that a common norm bound is not a
common intervention across models. RESCAL SafeSucc peaks at only $.013$
at $B=.3$ before damage dominates; on WN18RR, SafeSucc at $B=1$ ranges
from $.044$ to $.207$, with only $.143$--$.481$ of cases applicable,
meaning that enough eligible blockers exist to reach the top 10.
Norm budgets therefore require scorer-specific calibration.

\textbf{Available positive labels and confidence filtering.}
Table~\ref{tab:demotion-controls} identifies blocker selection, rather
than optimization alone, as a major safety mechanism. Removing
positive-label exclusion raises success to $.597/.608$ but collapses
the no-damage rate to $.192/.205$; increasing $q$ produces the same
trade-off. Demotion is safe only to the extent that its blocker set is
trustworthy.

\begin{table}[t]
\centering
\tablebodyfont
\caption{Gradient demotion across update bounds ($N=1{,}000$).}
\label{tab:norm-sensitivity}
\textit{(a) FB15k-237 RESCAL}\par\smallskip
\setlength{\tabcolsep}{1.25pt}
\begin{tabular}{rrrrrrr}
\toprule
$B$ & Success & \shortstack{No\\damage} & SafeSucc &
\shortstack{Mean\\$D$} & \shortstack{P95\\$D$} & \shortstack{Max.\\$D$} \\
\midrule
.01 & .003 & .871 & .000 & 4.783 & 9 & 786 \\
.03 & .011 & .845 & .001 & 11.239 & 35 & 1,655 \\
.10 & .087 & .823 & .011 & 30.527 & 168.55 & 1,945 \\
.30 & .210 & .792 & \textbf{.013} & 102.343 & 595.55 & 3,552 \\
1.0 & .228 & .773 & .001 & 134.402 & 1,002.75 & 3,662 \\
\bottomrule
\end{tabular}

\medskip
\textit{(b) WN18RR: Safe rate by update bound}\par\smallskip
\setlength{\tabcolsep}{1.8pt}
\begin{tabular}{lrrrrrr}
\toprule
Scorer & \shortstack{Appli-\\cable} & .01 & .03 & .10 & .30 & 1.0 \\
\midrule
DistMult & .290 & .000 & .000 & .000 & .006 & .044 \\
ComplEx  & .257 & .001 & .003 & .064 & .142 & .195 \\
RESCAL   & .223 & .000 & .001 & .006 & .042 & .138 \\
RotatE   & .143 & .000 & .004 & .008 & .017 & .054 \\
TransE   & .481 & .000 & .001 & .005 & .036 & .207 \\
\bottomrule
\end{tabular}

\medskip
\textit{(c) WN18RR damage at $B=1$}\par\smallskip
\setlength{\tabcolsep}{4.3pt}
\begin{tabular}{lrrr}
\toprule
Scorer & No damage & Mean $D$ & Max. $D$ \\
\midrule
DistMult & .951 & .107 & 9 \\
ComplEx  & .942 & .308 & 44 \\
RESCAL   & .964 & .088 & 9 \\
RotatE   & .970 & .074 & 10 \\
TransE   & .905 & .219 & 16 \\
\bottomrule
\end{tabular}
\end{table}

\begin{table}[t]
\centering
\tablebodyfont
\caption{Closed-form demotion across blocker-selection rules ($N=1{,}000$).}
\label{tab:demotion-controls}
\textit{(a) Positive labels used to exclude blockers}\par\smallskip
\setlength{\tabcolsep}{1.15pt}
\begin{tabular}{llrrrrr}
\toprule
Model & Labels & Success & \shortstack{No\\damage} &
SafeSucc & \shortstack{Mean\\$D$} & \shortstack{Norm\\ratio} \\
\midrule
DistMult & Train & .223 & .901 & .197 & .234 & .501 \\
 & Train+valid & .210 & .909 & .188 & .210 & .484 \\
 & All splits & .201 & .915 & .182 & .196 & .479 \\
 & No exclusion & .597 & .192 & .156 & 4.822 & 2.029 \\
\midrule
ComplEx & Train & .243 & .902 & .222 & .253 & .293 \\
 & Train+valid & .234 & .913 & .218 & .219 & .277 \\
 & All splits & .223 & .921 & .212 & .194 & .266 \\
 & No exclusion & .608 & .205 & .180 & 4.561 & .975 \\
\bottomrule
\end{tabular}

\medskip
\textit{(b) Confidence-filter sweep}\par\smallskip
\setlength{\tabcolsep}{3.8pt}
\begin{tabular}{rrrrr}
\toprule
& \multicolumn{2}{c}{DistMult} & \multicolumn{2}{c}{ComplEx} \\
\cmidrule(lr){2-3}\cmidrule(lr){4-5}
$q$ & Success & SafeSucc & Success & SafeSucc \\
\midrule
0   & .000 & .000 & .000 & .000 \\
.25 & .054 & .054 & .063 & .063 \\
.50 & .110 & .109 & .119 & .118 \\
.75 & .156 & .153 & .185 & .182 \\
1.0 & .201 & .182 & .223 & .212 \\
\bottomrule
\end{tabular}
\end{table}

\subsection{Population and Evaluation Sensitivity}

\textbf{Target difficulty and cutoff.}
Closed-form demotion is a near-cutoff strategy: its S@10 falls from
$.537/.578$ at ranks 11--20 to $0/.003$ beyond rank 100. Rank-truncated
S@10 remains $.876$--$.951$, so its performance is not explained by
easy targets alone. On rank-$>20$ cohorts, larger $k$ improves SafeSucc
for non-strict routes but barely changes strict success.

\textbf{Variation across relations.}
Relation composition materially changes the aggregate result. Entity
editing's case-bootstrap intervals, $[.333,.393]/[.346,.406]$, widen to
$[.252,.501]/[.279,.493]$ under relation clustering; equal relation
weighting raises SafeSucc to $.523/.517$. Case-weighted averages thus
partly reflect which relations dominate the cohort.

\textbf{Sensitivity overlap.}
Sensitivity overlap is a weak damage predictor: Pearson $r$ is only
$-.184$ ($[-.221,-.142]$) and $-.163$ ($[-.210,-.118]$). Local
geometric similarity does not summarize the resulting rank competition.

\textbf{Ranking convention and protected set.}
Filtered ranking raises direct and rank-truncated SafeSucc to
$.564/.547$ and $.790/.798$ by removing known positives from
competition. Protected-set size has a parallel effect: expanding from
500 to 5,000 to all training facts changes rank-truncated SafeSucc from
$.638/.328/.303$ and $.626/.347/.320$, with all-split protection lower
still at $.281/.284$. Safety estimates therefore improve mechanically
when the evaluation removes competitors or protects fewer answers; we
retain raw ranking and the stated training-only scope.

\subsection{External Case Study: KGEditor}
\label{sec:kgeditor}

On KGEditor's official 3,086-case FB15k-237 set, our reproduction
matches Hits@1/3 ($.8594/.9867$). Target success at 10 is $.9916$, but
$12.38\%$ of edits still displace at least one protected relation-scope
answer. Although the
unpaired setting prevents a controlled method comparison, the case study
shows that rank displacement persists even for a learned editor with very
high target success.

\section{Discussion and Limitations}
\textbf{Practical recommendation.} The protected scope should be chosen
before an editor is evaluated. Applications that return multiple answers
to one query require same-query locality; if the edited entity can also
compete across queries with the same relation, relation-scope locality is
additionally required. Parameter-support locality remains useful for
diagnosing direct interference, but should not be used alone as evidence
of rank preservation. In every setting, joint success and damage severity
should accompany correction success.

\textbf{Scope.} We study single edits to low-ranked held-out triples on
fixed checkpoints, not false-fact correction, unseen entities, or edit
sequences. Exact-preservation claims apply to the stated linear
parameterization and protected scores. Entity editing and demotion use
retrospective observed labels; deployment requires protection and
blocker-selection rules available at edit time. The primary comparison
uses FB15k-237 DistMult/ComplEx, with dimension, scorer, dataset, and
KGEditor extensions.

\section{Conclusion}
Correcting a target rank can displace correct answers while appearing
local under parameter support. Exact preservation rarely succeeds;
partial rank preservation improves joint success non-monotonically, a
gain absent from the ridge sweep. Reporting the protected scope, joint
success, and damage severity reveals both correction and side effects.

\bibliographystyle{IEEEtran}
\bibliography{references}

@misc{alphaedit,
      title={AlphaEdit: Null-Space Constrained Knowledge Editing for Language Models},
      author={Junfeng Fang and Houcheng Jiang and Kun Wang and Yunshan Ma and Shi Jie and Xiang Wang and Xiangnan He and Tat-seng Chua},
      year={2025},
      eprint={2410.02355},
      archivePrefix={arXiv},
      primaryClass={cs.CL},
      url={https://arxiv.org/abs/2410.02355}
}

@inproceedings{kgeditor,
      author = {Cheng, Siyuan and Zhang, Ningyu and Tian, Bozhong and Chen, Xi and Liu, Qingbin and Chen, Huajun},
      title = {Editing language model-based knowledge graph embeddings},
      year = {2024},
      isbn = {978-1-57735-887-9},
      publisher = {AAAI Press},
      url = {https://doi.org/10.1609/aaai.v38i16.29737},
      doi = {10.1609/aaai.v38i16.29737},
      booktitle = {Proceedings of the Thirty-Eighth AAAI Conference on Artificial Intelligence},
      series = {AAAI'24}
}

@inproceedings{ke,
      title = {Editing Factual Knowledge in Language Models},
      author = {De Cao, Nicola and Aziz, Wilker and Titov, Ivan},
      booktitle = {Proceedings of the 2021 Conference on Empirical Methods in Natural Language Processing},
      month = nov,
      year = {2021},
      address = {Online and Punta Cana, Dominican Republic},
      publisher = {Association for Computational Linguistics},
      url = {https://aclanthology.org/2021.emnlp-main.522/},
      doi = {10.18653/v1/2021.emnlp-main.522},
      pages = {6491--6506}
}

@inproceedings{mend,
      title={Fast Model Editing at Scale},
      author={Eric Mitchell and Charles Lin and Antoine Bosselut and Chelsea Finn and Christopher D. Manning},
      booktitle={International Conference on Learning Representations},
      year={2022},
      url={https://openreview.net/forum?id=0DcZxeWfOPt}
}

@inproceedings{fedlu,
      author = {Zhu, Xiangrong and Li, Guangyao and Hu, Wei},
      title = {Heterogeneous Federated Knowledge Graph Embedding Learning and Unlearning},
      year = {2023},
      isbn = {9781450394161},
      publisher = {Association for Computing Machinery},
      address = {New York, NY, USA},
      doi = {10.1145/3543507.3583305},
      booktitle = {Proceedings of the ACM Web Conference 2023},
      pages = {2444--2454},
      series = {WWW '23}
}

@article{temporal,
      title = {A survey on temporal knowledge graph embedding: Models and applications},
      journal = {Knowledge-Based Systems},
      volume = {304},
      pages = {112454},
      year = {2024},
      issn = {0950-7051},
      doi = {10.1016/j.knosys.2024.112454},
      author = {Yuchao Zhang and Xiangjie Kong and Zhehui Shen and Jianxin Li and Qiuhua Yi and Guojiang Shen and Bo Dong}
}

@inproceedings{transe,
      title={Translating Embeddings for Modeling Multi-relational Data},
      author={Bordes, Antoine and Usunier, Nicolas and Garcia-Dur{\'a}n, Alberto and Weston, Jason and Yakhnenko, Oksana},
      booktitle={Advances in Neural Information Processing Systems (NeurIPS)},
      volume={26},
      pages={2787--2795},
      year={2013}
}

@inproceedings{distmult,
      title={Embedding Entities and Relations for Learning and Inference in Knowledge Bases},
      author={Yang, Bishan and Yih, Wen-tau and He, Xiaodong and Gao, Jianfeng and Deng, Li},
      booktitle={International Conference on Learning Representations (ICLR)},
      year={2015}
}

@inproceedings{complex,
      title={Complex Embeddings for Simple Link Prediction},
      author={Trouillon, Th{\'e}o and Welbl, Johannes and Riedel, Sebastian and Gaussier, {\'E}ric and Bouchard, Guillaume},
      booktitle={International Conference on Machine Learning (ICML)},
      pages={2071--2080},
      year={2016}
}

@inproceedings{rescal,
      title={A Three-Way Model for Collective Learning on Multi-Relational Data},
      author={Nickel, Maximilian and Tresp, Volker and Kriegel, Hans-Peter},
      booktitle={International Conference on Machine Learning (ICML)},
      pages={809--816},
      year={2011}
}

@inproceedings{rotate,
      title={RotatE: Knowledge Graph Embedding by Relational Rotation in Complex Space},
      author={Sun, Zhiqing and Deng, Zhi-Hong and Nie, Jian-Yun and Tang, Jian},
      booktitle={International Conference on Learning Representations (ICLR)},
      year={2019}
}

@inproceedings{rome,
      title={Locating and Editing Factual Associations in {GPT}},
      author={Meng, Kevin and Bau, David and Andonian, Alex and Belinkov, Yonatan},
      booktitle={Advances in Neural Information Processing Systems (NeurIPS)},
      year={2022}
}

@inproceedings{memit,
      title={Mass-Editing Memory in a Transformer},
      author={Meng, Kevin and Sharma, Arnab Sen and Andonian, Alex and Belinkov, Yonatan and Bau, David},
      booktitle={International Conference on Learning Representations (ICLR)},
      year={2023}
}

@inproceedings{
gpm,
title={Gradient Projection Memory for Continual Learning},
author={Gobinda Saha and Isha Garg and Kaushik Roy},
booktitle={International Conference on Learning Representations},
year={2021},
url={https://openreview.net/forum?id=3AOj0RCNC2}
}

@inproceedings{ogd,
  title={Orthogonal gradient descent for continual learning},
  author={Farajtabar, Mehrdad and Azizan, Navid and Mott, Alex and Li, Ang},
  booktitle={International conference on artificial intelligence and statistics},
  pages={3762--3773},
  year={2020},
  organization={PMLR}
}

@inproceedings{
agem,
title={Efficient Lifelong Learning with A-{GEM}},
author={Arslan Chaudhry and Marc’Aurelio Ranzato and Marcus Rohrbach and Mohamed Elhoseiny},
booktitle={International Conference on Learning Representations},
year={2019}
}

@article{
ewc,
author = {James Kirkpatrick  and Razvan Pascanu  and Neil Rabinowitz  and Joel Veness  and Guillaume Desjardins  and Andrei A. Rusu  and Kieran Milan  and John Quan  and Tiago Ramalho  and Agnieszka Grabska-Barwinska  and Demis Hassabis  and Claudia Clopath  and Dharshan Kumaran  and Raia Hadsell },
title = {Overcoming catastrophic forgetting in neural networks},
journal = {Proceedings of the National Academy of Sciences},
volume = {114},
number = {13},
pages = {3521-3526},
year = {2017},
doi = {10.1073/pnas.1611835114},
eprint = {https://www.pnas.org/doi/pdf/10.1073/pnas.1611835114}}

@inproceedings{dicgrl,
    title = "{D}isentangle-based {C}ontinual {G}raph {R}epresentation {L}earning",
    author = "Kou, Xiaoyu  and
      Lin, Yankai  and
      Liu, Shaobo  and
      Li, Peng  and
      Zhou, Jie  and
      Zhang, Yan",
    editor = "Webber, Bonnie  and
      Cohn, Trevor  and
      He, Yulan  and
      Liu, Yang",
    booktitle = "Proceedings of the 2020 Conference on Empirical Methods in Natural Language Processing (EMNLP)",
    month = nov,
    year = "2020",
    address = "Online",
    publisher = "Association for Computational Linguistics",
    url = "https://aclanthology.org/2020.emnlp-main.237/",
    doi = "10.18653/v1/2020.emnlp-main.237",
    pages = "2961--2972"
}

@inproceedings{lkge,
  title = {Lifelong Embedding Learning and Transfer for Growing Knowledge Graphs},
  author = {Cui, Yuanning and 
            Wang, Yuxin and 
            Sun, Zequn and 
            Liu, Wenqiang and 
            Jiang, Yiqiao and 
            Han, Kexin and 
            Hu, Wei},
  booktitle = {AAAI},
  year = {2023}
}

@inproceedings{incde,
author = {Liu, Jiajun and Ke, Wenjun and Wang, Peng and Shang, Ziyu and Gao, Jinhua and Li, Guozheng and Ji, Ke and Liu, Yanhe},
title = {Towards continual knowledge graph embedding via incremental distillation},
year = {2024},
isbn = {978-1-57735-887-9},
publisher = {AAAI Press},
url = {https://doi.org/10.1609/aaai.v38i8.28722},
doi = {10.1609/aaai.v38i8.28722},
booktitle = {Proceedings of the Thirty-Eighth AAAI Conference on Artificial Intelligence and Thirty-Sixth Conference on Innovative Applications of Artificial Intelligence and Fourteenth Symposium on Educational Advances in Artificial Intelligence},
articleno = {974},
numpages = {10},
series = {AAAI'24/IAAI'24/EAAI'24}
}

@article{cohen-ripple,
  title = {Evaluating the Ripple Effects of Knowledge Editing in Language Models},
  author = {Cohen, Roi and Biran, Eden and Yoran, Ori and Globerson, Amir and Geva, Mor},
  journal = {Transactions of the Association for Computational Linguistics},
  volume = {12},
  pages = {283--298},
  year = {2024},
  doi = {10.1162/tacl_a_00644}
}

@inproceedings{li-pitfalls,
  title = {Unveiling the Pitfalls of Knowledge Editing for Large Language Models},
  author = {Li, Zhoubo and Zhang, Ningyu and Yao, Yunzhi and Wang, Mengru and Chen, Xi and Chen, Huajun},
  booktitle = {International Conference on Learning Representations},
  year = {2024},
  url = {https://openreview.net/forum?id=fNktD3ib16}
}

\end{document}